\documentclass[10pt,journal,final,a4paper,twoside,onecolumn]{IEEEtran}

\usepackage{slashbox}
\usepackage{amsmath}                        % Symboles mathématiques
\usepackage{graphicx}
\usepackage{amssymb}
\usepackage{algorithm}
\usepackage[noend]{algpseudocode} 
\usepackage{array}
\usepackage{stfloats}
\usepackage{float}
\usepackage[caption = false]{subfig}
\usepackage{multicol}
\usepackage[all,poly]{xy}
\usepackage{multirow}
\usepackage{color}
\usepackage{graphicx,psfrag}
\usepackage{mathtools, cuted,hyperref}

\makeatletter
\def\BState{\State\hskip-\ALG@thistlm}
\makeatother

\newlength{\tempdima}

\newcommand{\rowname}[1]% #1 = text
{\rotatebox{90}{\makebox[\tempdima][c]{#1}}}

\def\prox{{\textrm{prox}}}
\def\proj{{\textrm{Proj}}}

\def\bsOmg{{\boldsymbol{\Omega}}}
\def\bfb{{\mathbf{b}}}

\def\bfd{{\mathbf{d}}}

\def\bff{{\mathbf{f}}}

\def\bfk{{\mathbf{k}}}

\def\bfn{{\mathbf{n}}}

\def\bfr{{\mathbf{r}}}

\def\bfu{{\mathbf{u}}}
\def\bfv{{\mathbf{v}}}
\def\bfw{{\mathbf{w}}}
\def\bfx{{\mathbf{x}}}
\def\bfy{{\mathbf{y}}}
\def\bfz{{\mathbf{z}}}

\def\bfC{{\mathbf{C}}}

\def\bfI{{\mathbf{I}}}

\def\bfM{{\mathbf{M}}}

\def\calA{{\mathcal{A}}}

\def\calM{{\mathcal{M}}}

\def\calT{{\mathcal{T}}}

\newcommand{\rev}[1]{\textcolor[rgb]{0.00,0.00,0.00}{#1}}
\newcommand{\tbme}[1]{\textcolor[rgb]{0.00,0.00,0.00}{#1}}
\newcommand{\zhao}[1]{\textcolor[rgb]{0.00,0.00,0.00}{#1}}
\begin{document}
\bstctlcite{IEEEexample:BSTcontrol}
\title{\rev{Motion Compensated Dynamic MRI Reconstruction with Local Affine Optical Flow Estimation}}
%\title{Dynamic MRI Reconstruction based on Motion Estimation/Compensation with Local Affine Optical}
	
\author{Ningning Zhao, Daniel O'Connor, Adrian Basarab, Dan Ruan, Ke Sheng
\thanks{Ningning Zhao, Daniel O'Connor, Dan Ruan and Ke Sheng are with the department of radiation oncology, University of California, Los Angeles, CA, USA (e-mail: \{buaazhaonn, daniel.v.oconnor\}@gmail.com, \{DRuan, KSheng\}@mednet.ucla.edu). 

Adrian Basarab is with University of Toulouse, IRIT, CNRS UMR 5505, 118 Route de Narbonne, F-31062, Toulouse Cedex 9, France (e-mail: adrian.basarab@irit.fr).} 
}

\maketitle
\begin{abstract}
This paper proposes a novel framework to reconstruct the dynamic magnetic resonance images (DMRI) with motion compensation (MC). Specifically, by combining the intensity-based optical flow (OF) constraint with the traditional CS scheme, we are able to jointly reconstruct the DMRI sequences and estimate the inter frame motion vectors. Then, the DMRI reconstruction can be refined through MC with the estimated motion field. By employing the coarse-to-fine multi-scale resolution strategy, we are able to update the motion field in different spatial scales. The estimated motion vectors need to be interpolated to the finest resolution scale to compensate the DMRI reconstruction. Moreover, the proposed framework is capable of handling a wide class of prior information (regularizations) for DMRI reconstruction, such as sparsity, low rank and total variation. The formulated optimization problem is solved by a primal-dual algorithm with linesearch due to its efficiency when dealing with non-differentiable problems. Experiments on various DMRI datasets validate the reconstruction quality improvement using the proposed scheme in comparison to several state-of-the-art algorithms. 
\end{abstract}

\begin{keywords}  
Dynamic MRI, compressed sensing, optimization, primal-dual algorithm, line search, optical flow, multi-scale strategy, motion estimation/compensation 
\end{keywords}
%%%%%%%%%%%%%%%%%%%%%%%%%%%%%%%%%%%%%%%%%%%%%%%%%%%%%%%%%%%%%%%%%%%%%%%%%%%%%%%%%%%%%%%%%%%%%%%%%%%%
%%%%%%%%%%%%%%%%%%%%%%%%%%%%%%%%%%%%%%%%%%%%%%%%%%%%%%%%%%%%%%%%%%%%%%%%%%%%%%%%%%%%%%%%%%%%%%%%%%%%%
\section{Introduction}
\IEEEPARstart{D}{ynamic} magnetic resonance imaging (DMRI) plays an important role in different clinical exams, e.g., cardiovascular, pulmonary, abdominal, perfusion and functional imaging. The reconstruction of DMRI aims at obtaining spatio-temporal MRI sequences in $\bfx \mbox{-} t$ space, from their measurements acquired in the $\bfk \mbox{-} t$ space. The trade-off between spatial and temporal resolution in DMRI reconstruction is challenging due to the physical constraints. Classical techniques to deal with this issue include echo planar imaging \cite{Mansfield1977}, fast low-angle shot imaging \cite{HAASE2011} and parallel imaging \cite{Tsao2012}. 

In recent years, compressed sensing (CS) techniques have demonstrated great success in reducing the acquisition time without degrading image quality, see e.g.,\cite{Lustig2007,LustigMichaelandDonohoDavidLandSantosJuanMandPauly2008}. CS theory guarantees  \rev{an acceptable} recovery of specific signals or images from fewer measurements than the number predicted by the Nyquist limit. Image reconstruction from undersampled observations is an ill-posed problem that consequently requires prior information (regularization) to stabilize the solution. The regularizations widely used for DMRI reconstruction include sparsity in transformed domains \cite{Jung2009}, total variation (TV) penalties \cite{Knoll2012}, low-rank property \cite{Liang2007,Trzasko2011ISMRM,Miao2016} or a combination of several priors \cite{Lingala2011a,Majumdar2015}. Under the CS-based framework, DMRI reconstruction methods can be broadly divided into two categories: offline and online \cite{Majumdar2012}. Similar to most of CS-based DMRI reconstruction methods, we focus in this paper on the offline approach. 

Due to the presence of motion patterns in DMRI acquisition, combining the motion estimation/motion compenstaion (ME/MC) with the DMRI reconstruction has been explored in the literature, see e.g., \cite{Asif2013,Usman2013,Otazo2015,Tremoulheac2014,Lingala2015,Cordero-Grande2016,Cordero-Grande2017,Prieto2007,Jung2010,Feng2016,MRM2017Rank}. For instance, low rank plus sparse (L+S) matrix decomposition employed in DMRI reconstruction decomposes the DMRI sequences into two parts, where L models the temporally correlated background and S models the dynamic information \cite{Otazo2015,Tremoulheac2014}. Lingala \textit{et. al.} \cite{Lingala2015} coupled the DMRI reconstruction and the inter-frame motion estimation using a variable splitting algorithm. MaSTER algorithm \cite{Asif2013} was proposed to reconstruct DMRI followed by MC using motion vectors estimated with different strategies. In \cite{MRM2017Rank}, DMRI and motion estimation were conducted under multi-scale resolution framework. 

In this paper, we propose a novel DMRI reconstruction framework with MC, which includes two stages. One is variable updates, where the DMRI sequences and the inter-frame motion vectors are estimated jointly by combining an intensity-based optical flow (OF) constraint with the traditional CS scheme. In the second stage, the DMRI reconstruction is refined with the estimated motion vectors previously. By employing the coarse-to-fine multi-scale resolution strategy, we are able to estimate the motion vectors in different spatial resolution scales. The estimated motion vectors in a coarse scale are then interpolated to the finest scale in order to refine the image reconstruction. By varying the resolution scale, the two sub-problems are conducted alternately. Note that only the motion vectors are estimated in different resolution scales in the proposed algorithm, whereas both the image sequences and motion vectors were updated in different resolution scales in \cite{MRM2017Rank}. The formulated problems in the two stages are addressed using the primal-dual algorithm with linesearch \cite{Malitsky2016}, known to efficiently handle non-differentiable optimization problems. 

The contributions of this work are threefold: i) The primal dual algorithm with linesearch is explored to address the two sub-problems; ii) A wide class of DMRI priors can be handled in the general framework for jointly DMRI reconstruction and ME in the first stage; iii) In order to model local tissue deformations, an affine model is employed for the ME \cite{Suhling2005}. The proposed algorithm is an extension of our previous work \cite{nzhao20158SPIE}, where a reference frame is considered for ME. Experiments on three DMRI datasets demonstrate the superiority of the proposed framework over several state-of-the-art algorithms.

The remainder of this paper is organized as follows. In Section II, we describe the background related with the proposed framework. The variational problem is formulated in Section III. Section IV details the proposed algorithm. Section V gives the experimental results. Conclusions and perspectives are reported in Section VI. 
%%%%%%%%%%%%%%%%%%%%%%%%%%%%%%%%%%%%%%%%%%%%%%%%%%%%%%%%%%%%%%%%
%%%%%%%%%%%%%%%%%%%%%%%%%%%%%%%%%%%%%%%%%%%%%%%%%%%%%%%%%%%%%%%%
\section{Background}
In this section, the DMRI formation model is expressed. Moreover, the OF equation and its variants, the proximal operator and the primal-dual algorithm are illustrated hereinafter to facilitate the explanation of the proposed algorithm. 
\subsection{DMRI measurements}
The DMRI measurements acquired in the $\bfk \mbox{-} t$ space are denoted as $b_t(\bfk)$, 
which can be modelled by 
\begin{equation}
b_t(\bfk) = \int_{\bfx} f_t(\bfx) \exp(-j\bfk^T \bfx)d\bfx + n_t(\bfk)
\end{equation}
where $f_t(\bfx)$ \rev{of size $N_x\times N_y$} is the $t$th frame of the DMRI sequences, $n_t(\bfk)$ represents the additive white Gaussian noise, $\bfx = [x,y]\rev{^T}$ and $t$ are the spatial and temporal coordinates, $\bfk$ is the 2D frequency variable, $t\in\{1,\cdots,N_t\}$ with $N_t$ as the total number of temporal frames. Note that although the image formation model is valid for any number of spatial dimensions, to simplify the description, we only consider the $2D+t$ case in this paper \cite{MRI_handbook2004}. 
\rev{Given the matrix $\bff = [\bff_1,\cdots,\bff_{N_t}]$ of size $(N_x N_y)\times N_t$ whose column $\bff_t$ of size $N_xN_y\times1$ represents the vectorized version of the $t$th temporal frame $f_t(\bfx)$, we rewrite} the above expression in a matrix-vector form as below 
\begin{equation}
\bfb = \calA (\bff) + \bfn
\end{equation}
where the measurement operator $\calA$ represents the partial/ masked Fourier transform on specific sampling locations,  \rev{the observation $\bfb$ and additive noise $\bfn$ are vectors of size $N_b\times1$ where $N_b\ll ((N_x N_y)\times N_t)$.} 

\subsection{Optical flow}
Denoting \rev{$f_{t}(\bfx)$} as a fixed image acquired at time $t$, the brightness/intensity constancy in DMRI is formulated as 
\begin{equation}
f_t(\bfx) = f_{t_0}(\bfx-\bfd(\bfx,t)) 
\label{eq_BrightConst}
\end{equation}
where $\bfd(\bfx,t) = [\bfu(\bfx,t),\bfv(\bfx,t)]^T$ is the motion field between the fixed image and the moving frame \rev{$f_{t_0}(\bfx)$}, $\bfu(\bfx,t)$ and $\bfv(\bfx,t)$ are the horizontal and vertical components of the motion field. Under the hypothesis of small displacements, the first-order Taylor approximation can be used to replace the nonlinear intensity profile, \textit{i.e.}, 
\begin{equation}
f_{t_0}(\bfx-\bfd(\bfx,t)) \approx f_{t_0}- \partial_x f_{t_0} \bfu(\bfx,t) - \partial_y f_{t_0} \bfv(\bfx,t)
\label{TayerExpans}
\end{equation}
where the frame $f_{t_0}\triangleq f_{t_0}(\bfx)$, $\partial_xf_{t_0}$ and $\partial_yf_{t_0}$ are the partial derivatives of $f_{t_0}$ with respect to (w.r.t.) $x$ and $y$. Combining \eqref{eq_BrightConst} and \eqref{TayerExpans}, the traditional OF equation is given by
\begin{equation}
f_t(\bfx) - f_{t_0}+ \partial_x f_{t_0} \bfu(\bfx,t) + \partial_y f_{t_0} \bfv(\bfx,t) =0.
\label{eq_OFC}
\end{equation}
To estimate the motion vectors $\bfd(\bfx,t)$, a dedicated cost function can be formulated globally (on the entire image) or locally (by patches) using weighted OF \cite{SunCVPR2010,Altunbasak2003,Suhling2005,Alessandrini2013}.

\paragraph{Weighted OF \rev{and multiscale approach}}
The weighted OF equation can be expressed as below
\begin{equation}
\int_{\bfx} \bfw(\bfx-\bfx_0) \left[f_t(\bfx) - f_{t_0}+ \partial_x f_{t_0} \bfu(\bfx,t) + \partial_y f_{t_0} \bfv(\bfx,t)\right] d\bfx
\label{eq_WOF}
\end{equation}
where $\bfw$ is a window function centered at $\bfx_0$. Given the weighted OF equation, the motion vectors are assumed constant within a spatial neighbourhood. Moreover, B-spline based windows, \textit{i.e.}, $\bfw(\bfx)=\beta^n(x)\beta^n(y)$, where $\beta^n(\cdot)$ is a symmetrical B-spline function of degree $n\in\mathbb{N}$, have been shown to be adapted to medical images \cite{Suhling2005,Alessandrini2013}. The size of$\bfw$ is determined by the B-spline degree. 

Varying the resolution scale where the motion is estimated can be achieved by using a window function at different spatial scales. Specifically, the window function at spatial scale $j$ is expressed as below
\begin{equation}
\bfw^{(j)}(\bfx-\bfx_0) = \bfw\left(\frac{\bfx-2^j\bfx_0}{2^j}\right)
\label{eq_window}
\end{equation}
Since the window function at scale $j$ is dilated by a factor $2^j$, the calculation of \eqref{eq_WOF} at scale $j$ corresponds to subsampling of the inner product \eqref{eq_WOF} by a factor $2^j$. The coarse-to-fine multi-scale resolution approach has been demonstrated effective for myocardial motion estimation \cite{Suhling2005,Alessandrini2013}

\paragraph{Affine model}
It is important to note that the motion patterns in medical images can be very complex due to tissue deformations such as rotation, expansion, contraction and shear. In order to accurately describe these motion patterns, the affine model \rev{instead of the pure translation modelß} has been extensively used in the related literature, see e.g., \cite{Altunbasak2003,Suhling2005,Alessandrini2013}. Based on the affine model, the motion vectors at position $(x,y)$ for the $t$th frame are expressed by
\begin{equation}
\begin{array}{ll}
\bfu(\bfx,t) &= \bfu_0(\bfx,t) + \bfu_1(\bfx,t) x+ \bfu_2(\bfx,t) y \\
\bfv(\bfx,t) &= \bfv_0(\bfx,t) + \bfv_1(\bfx,t) x+ \bfv_2(\bfx,t) y
\end{array}
\label{eq_affine_model}
\end{equation}
where $\bfu_0$, $\bfu_1$, $\bfu_2$ and $\bfv_0$, $\bfv_1$, $\bfv_2$ are the 
affine parameters defining the motion of pixel at position $(x,y)$ in frame $t$ w.r.t. the reference frame $\bff_0$ \cite{Suhling2005}. 

\subsection{Proximal operator}
The proximal operator of a lower semicontinuous (l.s.c.) function $g$ is defined as  
\begin{equation}
\prox_{sg}(p) = \arg\min_{x} g(x) + \frac{1}{2s}\|x-p\|^2
\label{prox_opt}
\end{equation}
Note that the proximal operator calculation \eqref{prox_opt} always has a unique solution. 
One important property of the proximal operator is the Moreau's decomposition formula given by
\begin{equation}
\prox_{sg^*}(p) = p - s \prox_{s^{-1}g}\left(\frac{p}{s}\right).
\label{eq_Moreau}
\end{equation}
where $g^*$ is the convex conjugate of function $g$. Moreau's decomposition builds the relationship between the proximal operator of a l.s.c. function $g$ and the proximal operator of its conjugate \cite{OPT-003,pock2009algorithm}.  

\subsection{Primal-dual algorithm}
Primal-dual algorithms (PDAs) have been widely explored for non-smooth convex optimization problems, see e.g., \cite{Chambolle2011,Komodakis2015,pock2009algorithm,esser2009general}. Given an optimization problem as below
\begin{equation}
\min_{\bfy} g(\bfC\bfy) + h(\bfy)
\end{equation}
where $g$ and $h$ are proper, convex and l.s.c. functions, $\bfC$ is a continuous linear operator, the corresponding primal-dual/saddle-point problem is expressed by
\begin{equation}
\min_{\bfy}\max_{\bfz} \langle\bfC\bfy,\bfz \rangle + h(\bfy) - g^*(\bfz)
\label{primal_dual}
\end{equation}
where $\langle \cdot,\cdot\rangle$ is the inner product, $g^*$ is the conjugate of function $g$ and $\bfz$ is the dual variable. PDA seeks a solution $(\hat{\bfy},\hat{\bfz})$ of the problem \eqref{primal_dual} by alternating proximal gradient steps w.r.t. the primal and dual variables. 
%Algorithm \ref{algo_CP} summarizes the standard PDA, \rev{where $\bfC^*$ represents the adjoint of matrix $\bfC$.} Note that the stepsize parameters in PDA need to satisfy the relationship $s\sigma\|\bfC\|\leq 1$ to ensure the convergence. 
Different variants of PDA have been proposed more recently to tune the stepsize parameters adaptively and/or speed up the existing algorithms, see e.g., \cite{Komodakis2015, Malitsky2016}. 
Algorithm \ref{algo_CPL} summarizes the PDA with linesearch (PDAL), which accelerates the traditional PDA. 
$\bfC^*$ represents the adjoint of matrix $\bfC$.
%\begin{algorithm}
%\caption{Primal Dual Algorithm (PDA)}
%\label{algo_CP}
%\begin{algorithmic}[1]
%\Require{$\bfy^0$, $\bfz^0$, $\sigma$, $s$}
%\For{k = 1 $\cdots$}
%\State $\bfy^k = \prox_{\sigma h}(\bfy^{k-1} - \sigma\bfC^* \bfz^{k-1})$
%\State $\bfz^k = \prox_{sg^*}(\bfz^{k-1} + s\bfC(2\bfy^k - \bfy^{k-1}))$
%\State until stopping criterion is satisfied.
%\EndFor
%\end{algorithmic}
%\end{algorithm}
\begin{algorithm}
\caption{Primal Dual Algorithm with linesearch (PDAL)}
\label{algo_CPL}
\begin{algorithmic}[1]
\Require{$\bfy^0$, $\bfz^0$, $\sigma^0$, $s$, $\alpha>0$, $\epsilon\in(0,1)$, $\rho\in(0,1)$}
\State Set $\theta^0=1$.
\For{k = 1 $\cdots$}
\State $\bfy^k = \prox_{\sigma^{k-1} h}(\bfy^{k-1} - \sigma^{k-1}\bfC^* \bfz^{k-1})$
\State Choose any $\sigma^k \in [\sigma^{k-1},\sigma^{k-1}\sqrt{1+\theta^{k-1}}]$
\State \textit{\textbf{Linesearch}}
\State $\theta^k = \frac{\sigma^k}{\sigma^{k-1}}$
\State $\bar{\bfy}^k = \bfy^k + \theta^k(\bfy^k-\bfy^{k-1})$
\State $\bfz^k = \prox_{\alpha\sigma^k g^*}(\bfz^{k-1} + \alpha\sigma^k \bfC\bar{\bfy}^k)$
\If{$\sqrt{\alpha}\sigma^k \|\bfC^* \bfz^{k} - \bfC^* \bfz^{k-1} \| \leq \epsilon \|\bfz^{k} - \bfz^{k-1} \| $} 
\State Break linesearch
\Else
\State $\sigma^k = \sigma^k \rho$ and go to \textit{\textbf{linesearch}} (step 5)
\EndIf
\State Until stopping criterion is satisfied. 
\EndFor
\end{algorithmic}
\end{algorithm}

%%%%%%%%%%%%%%%%%%%%%%%%%%%%%%%%%%%%%%%%%%%%%%%%%%%%%%%%%%%%%%%
%%%%%%%%%%%%%%%%%%%%%%%%%%%%%%%%%%%%%%%%%%%%%%%%%%%%%%%%%%%%%%%
\section{Problem formulation}
The problem can be divided into two stages, which are detailed in this section.
\subsection{Joint DMRI reconstruction and motion estimation}
Given the matrix $\bar{\bff} = [\bff_{N_t},\bff_1,\cdots,\bff_{N_t-1}]$, i.e., $\bar{\bff}$ is $\bff$ with forward temporal shift by 1, the problem to joint reconstruct the DMRI and estimate the motion field at resolution scale $j$ is formulated by the following variational framework 
\begin{equation}
\min_{\bff,\bfd} \|\mathcal{A}(\bff)-\bfb\|^2  + \rev{\eta\phi(\calT\bff) + \tau\|\calM_{\bfw^{(j)}}(\bff,\bar{\bff},\bfd)\|_1} +\rev{\gamma} \psi(\bfd),   
\label{fomulated_problem}
\end{equation}
where $\phi(\calT \bff)$ is the regularization term incorporating prior information about the DMRI, $\calT$ represents a given transform, \rev{$\calM_{\bfw^{(j)}}(\bff,\bar{\bff},\bfd)$ is the weighted OF constraint between image sequences $\bff$ and $\bar{\bff}$ expressed in \eqref{eq_OFj}, $\bfd = [\bfu,\bfv]$ is the displacement field between $\bff$ and $\bar{\bff}$}, $\psi(\bfd)$ is a regularization term to smooth the displacement fields and $\eta$, $\tau$ and $\gamma$ are hyperparameters weighting the importance of each term. 
\begin{align}
&\rev{\calM_{\bfw^{(j)}}(\bff,\bar{\bff},\bfd)} \notag \\
= & \langle\bff - \bar{\bff} \rangle_{\bfw^{(j)}} + \langle\partial_x\bar{\bff}\rangle_{\bfw^{(j)}}\bfu +  \langle\partial_y\bar{\bff}\rangle_{\bfw^{(j)}} \bfv  \notag\\
= & \langle\bff - \bar{\bff} \rangle_{\bfw^{(j)}}+  \langle \partial_x\bar{\bff}\rangle_{\bfw^{(j)}}\bfu_0+  \langle x\partial_x\bar{\bff} \rangle_{\bfw^{(j)}}\bfu_1 +  \langle y\partial_x\bar{\bff}\rangle_{\bfw^{(j)}}\bfu_2   \notag\\ 
&+  \langle\partial_y\bar{\bff}\rangle_{\bfw^{(j)}} \bfv_0 
+\langle x\partial_y\bar{\bff}\rangle_{\bfw^{(j)}}\bfv_1 
+ \langle y\partial_y\bar{\bff}\rangle_{\bfw^{(j)}} \bfv_2
\label{eq_OFj}
\end{align}
where $\langle\bfr\rangle_{\bfw^{(j)}}$ is the weighted average of variable $\bfr \in \{\bff - \bar{\bff},\partial_x\bar{\bff},x\partial_x\bar{\bff},y\partial_x\bar{\bff},\partial_y\bar{\bff},x\partial_y\bar{\bff},y\partial_y\bar{\bff} \}$ at scale $j$, which is given by 
\begin{equation}
\langle\bfr\rangle_{\bfw^{(j)}} = \int_{\bfx} \bfw^{(j)}(\bfx-\bfx_0) \bfr(\bfx)d\bfx.
\label{eq_InnerProduct}
\end{equation}
%In this paper, the window function is constructed with B-spline functions as shown in Section II-B. 
%The calculation of \eqref{eq_InnerProduct} is related to the multi-resolution theory. Further details can be found in  \cite{Suhling2004,Suhling2005}. 

In order to smooth the displacement fields, the TV prior is used to regularize the motion vectors. Considering anisotropic TV, we have 
\begin{equation}
\psi(\bfd) =\sum_{i=0}^2 \|\nabla \bfu_i\|_1+ \sum_{i=0}^2 \|\nabla \bfv_i\|_1
\end{equation}
where
\begin{equation}
\|\nabla \cdot \|_1 = \sum_{i,j} \big \rvert (\nabla_x \cdot)_{i,j} \big \rvert  + \big \rvert  (\nabla_y \cdot)_{i,j}  \big \rvert 
\end{equation}
with 
\begin{equation}
(\nabla_x \cdot)_{i,j} = \left\{
                \begin{array}{ll}
                (\cdot)_{i+1,j} - (\cdot)_{i,j} & \textrm{if} \;\; i<N_x\\
                0 & \textrm{if} \;\; i=N_x
                \end{array}
                \right.
\end{equation}
\begin{equation}
(\nabla_y \cdot)_{i,j} = \left\{
                \begin{array}{ll}
                (\cdot)_{i,j+1} - (\cdot)_{i,j} & \textrm{if} \;\; j<N_y\\
                0 & \textrm{if} \;\; i=N_y
                \end{array}
                \right.
\end{equation}
\zhao{Note that $\ell_2$-norm prior can also be implemented to smooth the motion field since the proposed algorithm can easily handle a wide range of priors for the variables to be estimated.}
\subsection{Refining DMRI reconstruction by MC}
\rev{The inter-frame motion vectors estimated at spatial resolution $j$ are interpolated to the finest scale (the same as the image resolution scale). We then refine the reconstructed DMRI sequences by solving the following optimization problem.
\begin{equation}
\min_{\bff} \sum_t\|\mathcal{A}_t(\bff_t)-\bfb_t\|^2 + \lambda \|\bfM_ {t-1}\bff_{t-1} - \bff_t\|_1,
\label{eq_MC}
\end{equation}
where $\bff_t$ is the $t$th temporal frame of DMRI and $\bfM_{t-1}$ is the motion operator that uses the motion vectors to interpolate the pixels in MRI frame $\bff_{t-1}$ to displaced locations in $\bff_{t}$ \cite{Asif2013}.}
%%%%%%%%%%%%%%%%%%%%%%%%%%%%%%%%%%%%%%%%%%%%%%%%%%%%%%%%%%%%%%%
%%%%%%%%%%%%%%%%%%%%%%%%%%%%%%%%%%%%%%%%%%%%%%%%%%%%%%%%%%%%%%%
\section{Proposed algorithm}
\tbme{Note that both the formulated sub-problems can be solved using primal-dual algorithm. Hereinafter, we summarize the proposed algorithm.}
\subsection{\rev{Joint DMRI reconstruction and motion estimation}}
Since the formulated problem \eqref{fomulated_problem} is non-differentiable, we propose in this work a PDA-based algorithm to solve it. We first rewrite \eqref{fomulated_problem} as a sum of several l.s.c. functions as below
\begin{equation}
\min_{\bfy} g(\bfC\bfy) =  \sum_{l=1}^{9} g_l(\bfC_l \bfy) \triangleq  \sum_{l=1}^{9} g_l(\bsOmg_l)
\label{fomulated_problem_v2} 
\end{equation}
where $\bsOmg_l = \bfC_l \bfy$, $\bfy = [\bff,\bfu_0,\bfu_1,\bfu_2,\bfv_0,\bfv_1,\bfv_2]^T$ is the variable to be estimated, the matrix $\bfC$ is expressed in \eqref{matrixC} and the expression of functions $g_l(\cdot)$ ($l=1\cdots 9$) are expressed in \eqref{funcsG}.
%\newpage
%\begin{strip}
\begin{equation}
\bfC =
\begin{bmatrix}
\bfC_1 \\
\bfC_2 \\
\bfC_3 \\
\bfC_4 \\
\bfC_5 \\
\bfC_6 \\
\bfC_7 \\
\bfC_8 \\
\bfC_9 \\
%\bfC_{10} \\
\end{bmatrix}
= 
\begin{bmatrix}
\calA & 0 & 0& 0 &0 & 0 &0\\
\calT & 0 & 0& 0 &0 & 0 &0 \\
\langle\cdot\rangle_{\bfw^{(j)}} & \langle\partial_x\bar{\bff}\rangle_{\bfw^{(j)}} & \langle x\partial_x\bar{\bff}\rangle_{\bfw^{(j)}}&\langle y\partial_x\bar{\bff}\rangle_{\bfw^{(j)}}& \langle\partial_y \bar{\bff} \rangle_{\bfw^{(j)}}&\langle x\partial_y \bar{\bff} \rangle_{\bfw^{(j)}}&\langle y\partial_y \bar{\bff} \rangle_{\bfw^{(j)}} \\
0 & \nabla &0 &0 &0 &0 & 0 \\
0 & 0 &\nabla &0 &0 &0 & 0 \\
0 & 0 &0 &\nabla &0 &0 & 0 \\
0 & 0 &0 &0&\nabla &0 &0  \\
0 & 0 &0 &0&0 &\nabla &0  \\
0 & 0 &0 &0&0 &0 &\nabla  \\
\end{bmatrix},
\label{matrixC}
\end{equation}
%\end{strip}
\begin{equation}
\begin{cases}
g_1(\bsOmg_1) = \frac{1}{2}\|\bsOmg_1-\bfb\|^2, \\
g_2(\bsOmg_2) = \eta \phi(\bsOmg_2),    \\
g_3(\bsOmg_3) = \tau \|\bsOmg_3-\langle \bar{\bff}\rangle_{\bfw^{(j)}}\|_1, \\
g_l(\bsOmg_d) = \gamma \|\bsOmg_l\|_1, \;\textrm{for}\; \rev{l}=4,\ldots,9.  \\
\end{cases}
\label{funcsG}
\end{equation}
By introducing the dual variables $\bfz = [\bfz_1,\ldots,\bfz_{9}]^T$, the PDA iteration for problem \eqref{fomulated_problem_v2} can be summarized as follows 
\begin{equation}
\begin{array}{l}
\text{For}\; k = 0,\ldots,\\
\left\lfloor
\begin{array}{l}
\bfy^k = \bfy^{k-1} - \sigma \left(\sum_{l=1}^{9} \bfC_l^*\bfz^{k-1}_l\right),\\
\bfz^k_l =\prox_{sg^*_l}(\tilde{\bfz}^{k-1}_l), \\
\;\;\;\;\;= \prox_{sg^*_l}(\bfz^{k-1}_l + s\bfC_l(2\bfy^k - \bfy^{k-1})), 
\end{array}
\right.
\end{array}
\label{algo_PDA}
\end{equation}
where $\bfC^*_l$ is the adjoint of the matrix $\bfC_l$. The derivation of $\prox_{s g_2^*}(\cdot)$ is related to the expression of DMRI regularization functions.
The calculation of the rest proximal operator of $g_l^*$ $(l\neq 2)$ is given as below
\begin{equation}
\begin{cases}
\prox_{s g_1^*}(\tilde{\bfz}_1) = \frac{\tilde{\bfz}_1 -s\bfb}{1+s}, \\
\prox_{s g_3^*}(\tilde{\bfz}_3) = \proj_{\tau P}\left(\tilde{\bfz}_3-s\langle\bar{\bfI}_0\rangle_{\bfw^{(j)}}\right),\\
%\prox_{s g_3^*}(\tilde{\bfz}_3) = \prox_{s g_3^*}(\tilde{\bfz}_3),           \\
\prox_{s g_l^*}(\tilde{\bfz}_d) = \proj_{\gamma P}(\tilde{\bfz}_l), \;\textrm{for}\;l=4,\ldots,9,  \\
%\prox_{s g_{10}^*}(\tilde{\bfz}_{10}) = \proj_{\lambda P}(\tilde{\bfz}_{10}),
\end{cases}
\label{proxgstar}
\end{equation}
where $\proj_{\tau P}$ is a projector onto the convex set (Euclidean $\ell^2$-ball) $\tau P=\{\|p\|_{\infty}\leq \tau\}$, where $\|p\|_{\infty} = \max_{i,j}|p_{i,j}|$. In practice, this projector can be computed using the straightforward formula
\begin{equation}
\proj_{\tau P} (p) = \frac{p}{\max\{\tau,|p| \}}.
\label{prox_indicator}
\end{equation}

\begin{algorithm}
\caption{Joint MRI reconstruction and motion estimation using PDAL (JPDAL)}
\label{algo_ktPDAL}
\begin{algorithmic}[1]
\Require{$\bfy^0=[\bff^0,\bfu_0^0,\bfu_1^0,\bfu_2^0,\bfv_0^0,\bfv_1^0,\bfv_2^0]$, $\bfz_l^0$, $l\in\{1\cdots9\}$, $\sigma^0>0$, $\alpha>0$, $\epsilon\in(0,1)$, $\rho\in(0,1)$}
\State Set $\theta^0 = 1$
\For{k = 1 $\ldots$}
\Comment{Update $\bfy = [\bff,\bfu_0,\bfu_1,\bfu_2,\bfv_0,\bfv_1,\bfv_2]$}
\State $\bfy^k   = \bfy^{k-1}   - \sigma^{k-1} \left(\sum_{l=1}^{9} \bfC_l^*\bfz^{k-1}_l\right)$ 
\State Choose any $\sigma^k \in [\sigma^{k-1},\sigma^{k-1}\sqrt{1+\theta^{k-1}}]$
\State \textit{\textbf{Linesearch}} %\Comment{Start linesearch}
\State $\bar{\bfy}^k = \bfy^k + \theta^{k} (\bfy^k - \bfy^{k-1})$
\For{l=1, $\ldots$, 9}
\State $\bfz_l^{k} = \prox_{\alpha\sigma^k g_l^*}(\bfz_l^{k-1} + s\bfC_l \bar{\bfy}^k)$ %\Comment{Update $\bfz_l$ by calculating the proximal operator of $g_l^*(\cdot)$}
\EndFor 
\If{$\sqrt{\alpha}\sigma^k \|\bfC^T \bfz^{k} - \bfC^T \bfz^{k-1} \| \leq \epsilon \|\bfz^{k} - \bfz^{k-1} \| $} 
\State break the linesearch   %\Comment{Break linesearch}
\Else
\State $\sigma^k = \sigma^k \rho$ and go to \textit{\textbf{linesearch}} 
\EndIf
\State \rev{$\bar{\bff} = [\hat{\bff}_{N_t},\hat{\bff}_1,\cdots,\hat{\bff}_{N_t-1}]$} %\Comment{\rev{Update the matrix $\bar{\bfI}$ with the estimated image sequences $\hat{\bfI}$}}
\State Until stopping criterion is satisfied. %\Comment{Stopping criterion}
\EndFor
\end{algorithmic}
\end{algorithm}

\rev{In order to speed up \eqref{algo_PDA}, a variant of PDA with linesearch \cite{Malitsky2016} is employed. The resulting algorithm for jointly reconstructing DMRI and estimating the motion vectors at spatial scale $j$,} denoted as (JPDAL), is summarized in Algorithm \ref{algo_ktPDAL}. The stopping criterion employed is given by
\begin{equation}
\frac{\rvert L(\bfy^{k+1}) - L(\bfy^{k})\rvert}{L(\bfy^{k})} < \epsilon
\end{equation} 
where $L(\bfy)$ is the cost function. The stopping tolerance $\epsilon = 10^{-4}$ in this paper. 

\subsection{Proposed algorithm}
The proposed motion compensated DMRI reconstruction framework is summarized in Algorithm \ref{algo_proposed}, denoted as MC-JPDAL. \zhao{The proposed method alternates between two steps. In the first step, the MRI images and the inter-frame motion vectors (at specific resolution scale) are estimated jointly. Since the image sequences are estimated at the finest resolution scale, the estimated vectors are interpolated into the finest scale for the MC, i.e., the refinement of MRI reconstruction. }
In this paper, the range of the resolution scales where the motion vectors are estimated is fixed at $[J_c:J_f]$ with $J_c = 5$ and $J_f = 3$. \rev{The parameters of the proposed algorithm are divided into two groups. One group includes the parameters related to the PDAL, such as the step-size. They were fixed to $\sigma^0 = 1$, $\alpha = 0.5$, $\epsilon = 0.99$ \cite{Malitsky2016}. The second category composes the regularization parameters. In this paper, the regularization parameters $\eta$ and $\tau$ are tuned one-by-one in terms of quality of the reconstructed MRI by cross validation. In addition, the regularization terms for different dataset are chosen according to the reconstruction quality in this paper.}

\begin{algorithm}
\caption{\rev{Multi-scale Motion Compensated DMRI reconstruction using JPDAL (MC-JPDAL)}}
\label{algo_proposed}
\begin{algorithmic}[1]
\For{$j = J_c:J_f$} %\Comment{$[J_c,J_f]$ represents the range of the resolution scale}
\State \textbf{Variable estimation}: Solving \eqref{fomulated_problem} using \zhao{Algorithm 2};
\Comment{\zhao{Joint motion estimation and DMRI reconstruction.}}
\State \textbf{MC}: Solving \eqref{eq_MC} using \zhao{Algorithm 1}. 
\EndFor
\end{algorithmic}
\end{algorithm}
%%%%%%%%%%%%%%%%%%%%%%%%%%%%%%%%%%%%%%%%%%%%%%%%%%%%%%%%%%%%%%%
%%%%%%%%%%%%%%%%%%%%%%%%%%%%%%%%%%%%%%%%%%%%%%%%%%%%%%%%%%%%%%%
\section{Experimental results}
In order to evaluate the performance of the proposed algorithm, three MRI datasets were employed in this section: i) coronal lung image, ii) short-axis cardiac cine \footnote{The data was downloaded using the link \url{https://github.com/js3611/Deep-MRI-Reconstruction/tree/master/data}} and iii) two-chamber cardiac cine \footnote{The data was downloaded using the link \url{http://www.ece.ucr.edu/~sasif/dynamicMRI/index.html}}. \tbme{All three datasets were collected as fully-sampled data and retrospectively undersampled from single or multiple receiver coils according to a desired sampling pattern.}

A comparison between the proposed MC-JPDAL and different state-of-the-art algorithms, including ktSLR \cite{Lingala2011a}, L+S \cite{Otazo2015} and MaSTER \cite{Asif2013} was conducted in terms of the image reconstruction quality. The quantitative performance of different algorithms was evaluated using the root mean square error (RMSE) and the image structure similarity index (SSIM)~\cite{Wang2004}. The two metrics are expressed as below  
\begin{align}
\textrm{RMSE} &= \sqrt{E(\|\hat{\bff} - \bff\|_2^2)} \\
\textrm{SSIM} &= \frac{(2\mu_{\hat{\bff}}\mu_{\bff} + c_1)(2\sigma_{\hat{\bff}\bff} + c_2)}{(\mu_{\hat{\bff}}^2 + \mu_{\bff}^2 +c_1)(\sigma_{\hat{\bff}}^2 + \sigma_{\bff}^2 +c_2)}
\end{align}
where $\bff$, $\hat{\bff}$ are the ground truth and the estimated MRI sequences respectively, $E(\cdot)$ is the arithmetic mean, $\mu_a$ and $\sigma_a^2$ are the average and variance of variable $a$ ($a \in \{\hat{\bff}, \bff\}$), $\sigma_{\hat{\bff} \bff}$ is the covariance between $\hat{\bff}$ and $\bff$, $c_1$ and $c_2$ are two constants to stabilize the division with small denominator.   

\tbme{In order to evaluate how much each stage in MC-JPDAL contributes to the final reconstruction quality, we also compared the DMRI reconstruction performance using JPDAL and MC-JPDAL.} The initial guess of all the algorithms implemented in this paper was chosen by $\bff^0 = \mathcal{A}^T(\bfb)$. Experiments in this section were performed using MATLAB 2017b on a 64 bit Linux platform with Intel(R) Core(TM) i7-6700K CPU @4.00GHz and 48 GB RAM. 
\subsection{Coronal lung data}
The coronal lung data was acquired with a 1.5T Siemens Sonata Vision using spin echo (SE) sequences. The coronal lung data is of size $192\times 192 \times 40$ with pixel-size $2.08\times 2.08$ mm per frame and 40 temporal frames. The slice thickness is 7 mm. In this experiment, a golden angle radial sampling pattern \cite{Feng2014} was implemented. 

\begin{figure}[!h]
\begin{center}
\includegraphics[width=0.6\linewidth]{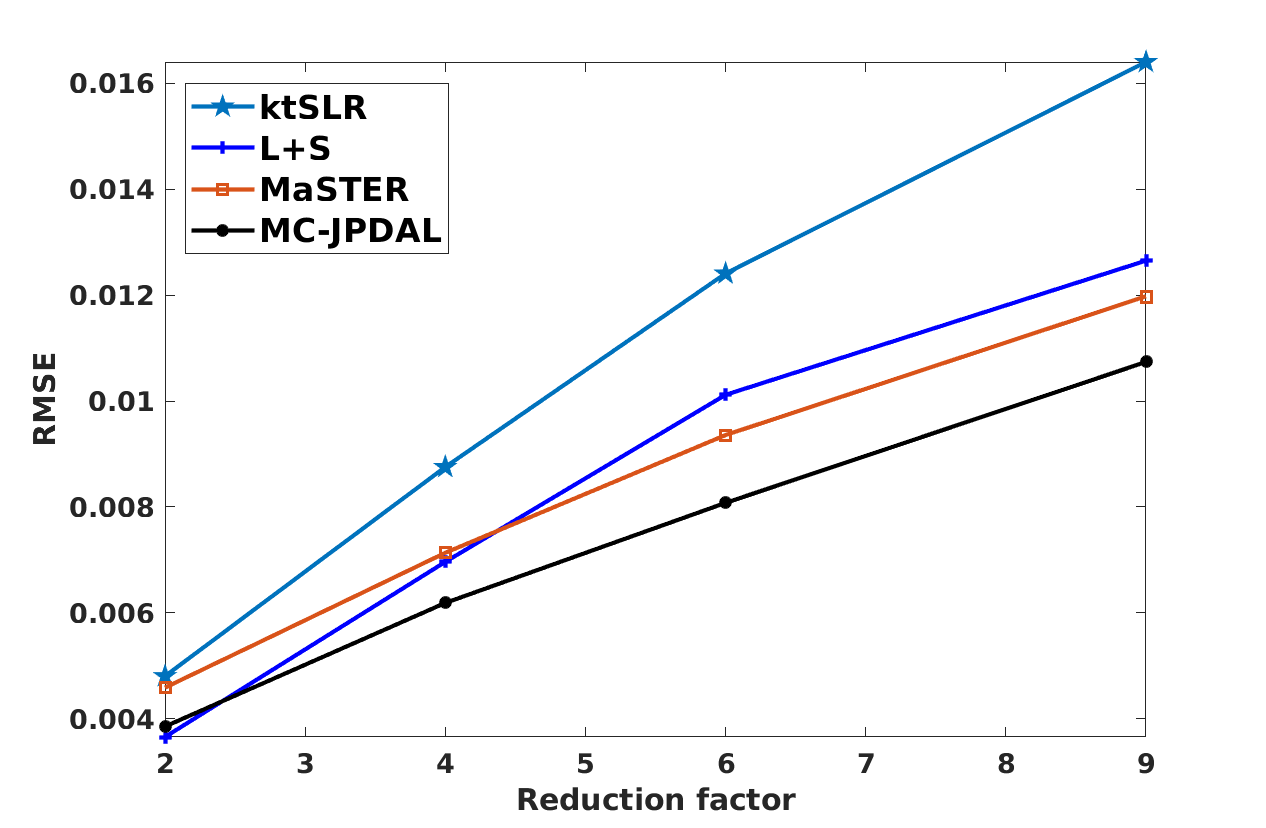}
\caption{RMSE comparison using different reduction factors for the coronal lung data with algorithms ktSLR, L+S, MaSTER and MC-JPDAL.}
\label{fig_CL_RD}
\end{center}
\end{figure}

Fig. \ref{fig_CL_RD} displays the reconstruction comparison with different reduction factors for the coronal lung data using algorithms ktSLR, L+S, MaSTER and the proposed MC-JPDAL. We observed that the proposed algorithm is superior to the others at different reduction scales in terms of RMSE. 

\vspace{-1cm}
\begin{figure}[!h]
\begin{center}
\includegraphics[width=0.6\linewidth]{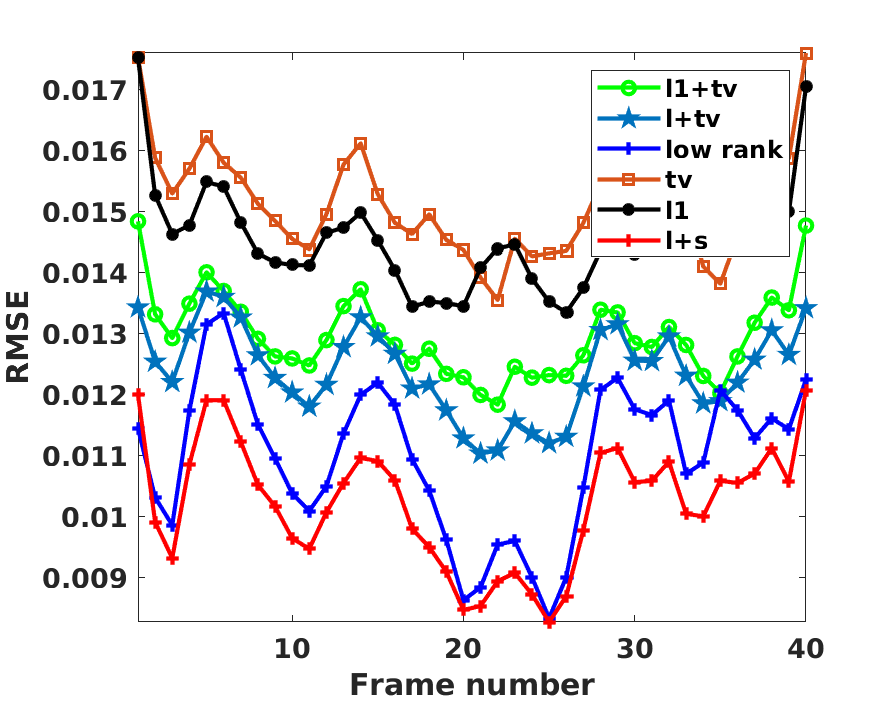}
\caption{RMSE comparison for the coronal lung MRI dataset using the proposed JPDAL with different priors: ``$\ell_1$+tv" (sparsity plus TV), ``l+tv" (low rank plus TV), ``tv" (TV), ``$\ell_1$" (sparsity), ``l+s" (low rank plus sparsity).}
\label{fig_CL_priors}
\end{center}
\end{figure}

\tbme{Fig. \ref{fig_CL_priors} shows the reconstruction comparison of the proposed JPDAL using different priors w.r.t. RMSE and SSIM. The reconstruction with prior ``l+s" (low rank and sparsity in temporal domain) outperforms the others according to Fig. \ref{fig_CL_priors}. Thus, the regularization term for the coronal lung dataset is chosen as ``l+s" in the proposed algorithms for further comparison.}

\begin{figure*}[!h]
\begin{center}
\includegraphics[width=0.95\linewidth]{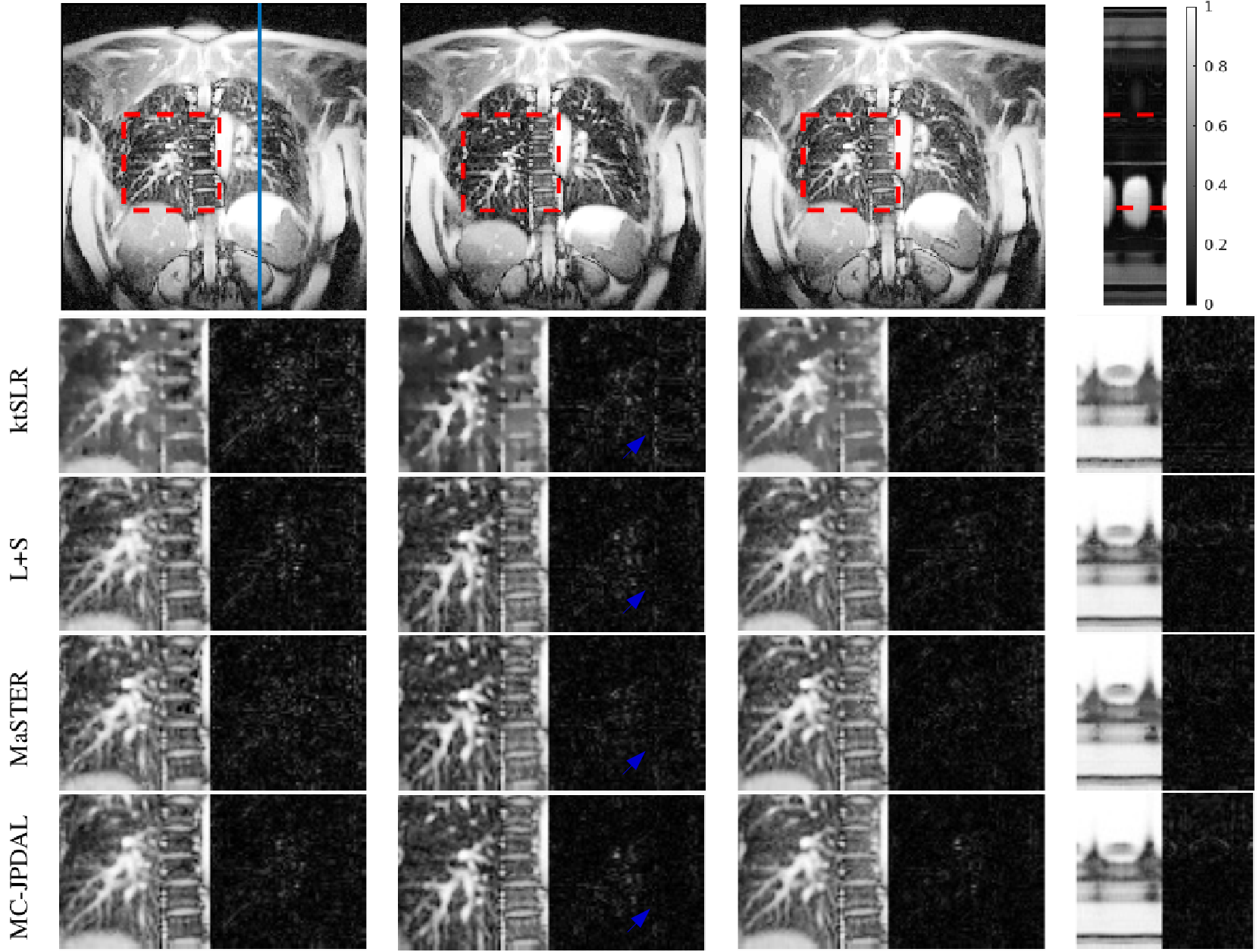}
\caption{Reconstruction of the coronal lung MRI scan using different algorithms: frame 1, 10 and 19 and the temporal profile (left to right). Top row: fully sampled MRI sequence with ROI contoured using red dashed rectangle and the location of the extracted temporal profile indicated using blue vertical line. Bottom rows: zoomed spatial ROI of the reconstructed MRI scans using ktSLR, L+S, MaSTER and the proposed MC-JPDAL.}
\label{fig_CL_MRI}
\end{center}
\end{figure*}

Fig. \ref{fig_CL_MRI} includes three example frames and the temporal profiles of the reconstructed DMRI using different algorithms at reduction factor 9. The first row shows the fully sampled coronal lung data at temporal frames 1, 10 and 19 and the temporal profile in $y$-$t$ space (from left to right). The location where the temporal profile extracted is indicated using a blue vertical line. The region of interest (ROI) are contoured using a red dashed rectangle. The zoomed ROIs and their corresponding difference images (i.e., $\bff - \hat{\bff}$) of the reconstructed MRI frames using algorithms ktSLR, L+S, MaSTER, and MC-JPDAL are displayed from 2nd to 5th rows. According to Fig. \ref{fig_CL_MRI}, \tbme{the magnitudes of the difference images obtained with the proposed algorithm MC-JPDAL is darker than the others.}

\begin{figure*}[!h]
\begin{center}
\includegraphics[width=0.95\linewidth]{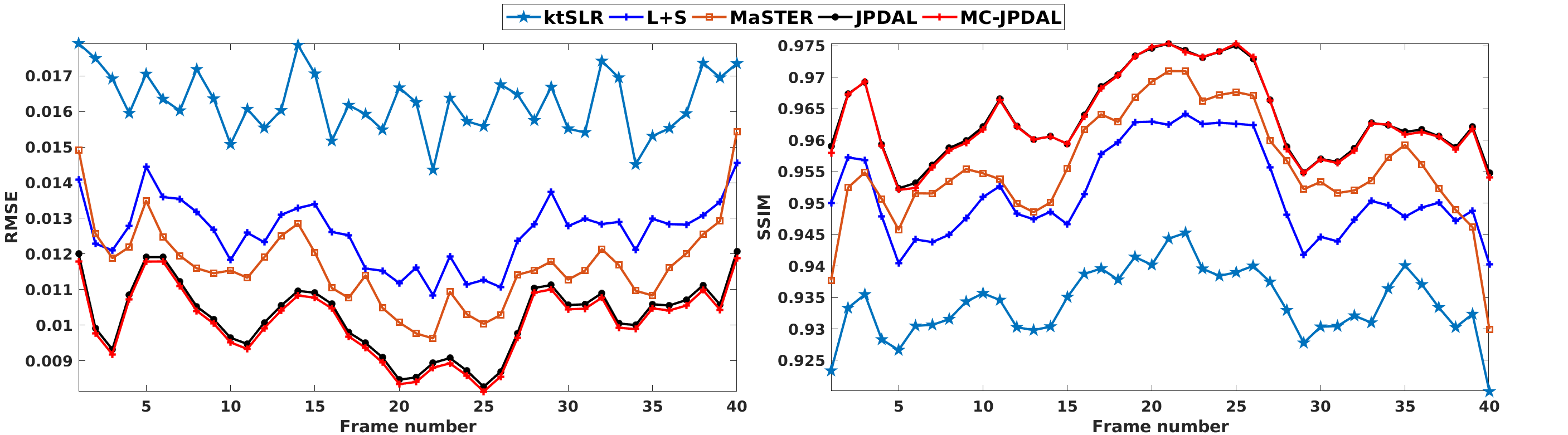}
\caption{Quantitative comparison of the lung coronal MRI sequences using the algorithms: ktSLR, L+S, MaSTER, \tbme{the proposed JPDAL and MC-JPDAL}. Left: RMSEs over the whole image; Right: SSIMs over the whole image.}
\label{fig_CL_SER}
\end{center}
\end{figure*}

The quantitative measurements calculated over the whole MRI frames are displayed in Fig. \ref{fig_CL_SER}. The proposed algorithm is superior to other algorithms in terms of the two RMSE and SSIM, which is consistent with the visual inspection. \tbme{We also observe that MC-JPDAL improves the DMRI reconstruction quality slightly comparing with JPDAL in Fig. \ref{fig_CL_SER}. }


\subsection{Short-axis cardiac cine data}
The cardiac cine data was used in \cite{Caballero2014}, which is of size $256\times 256$ per frame and contains 30 temporal frames. In this simulation, a golden angle radial down-sampling pattern with 24 rays per frame was performed. The corresponding down-sampling factor is 12. \tbme{After comparing different priors for the reconstruction of the cardiac cine MRI sequences, the prior for this dataset is the combination of sparsity and TV prior (denoted as ``$\ell_1$+tv") in the proposed algorithms for further comparison.}

\begin{figure*}[!h]
\begin{center}
\includegraphics[width=0.95\linewidth]{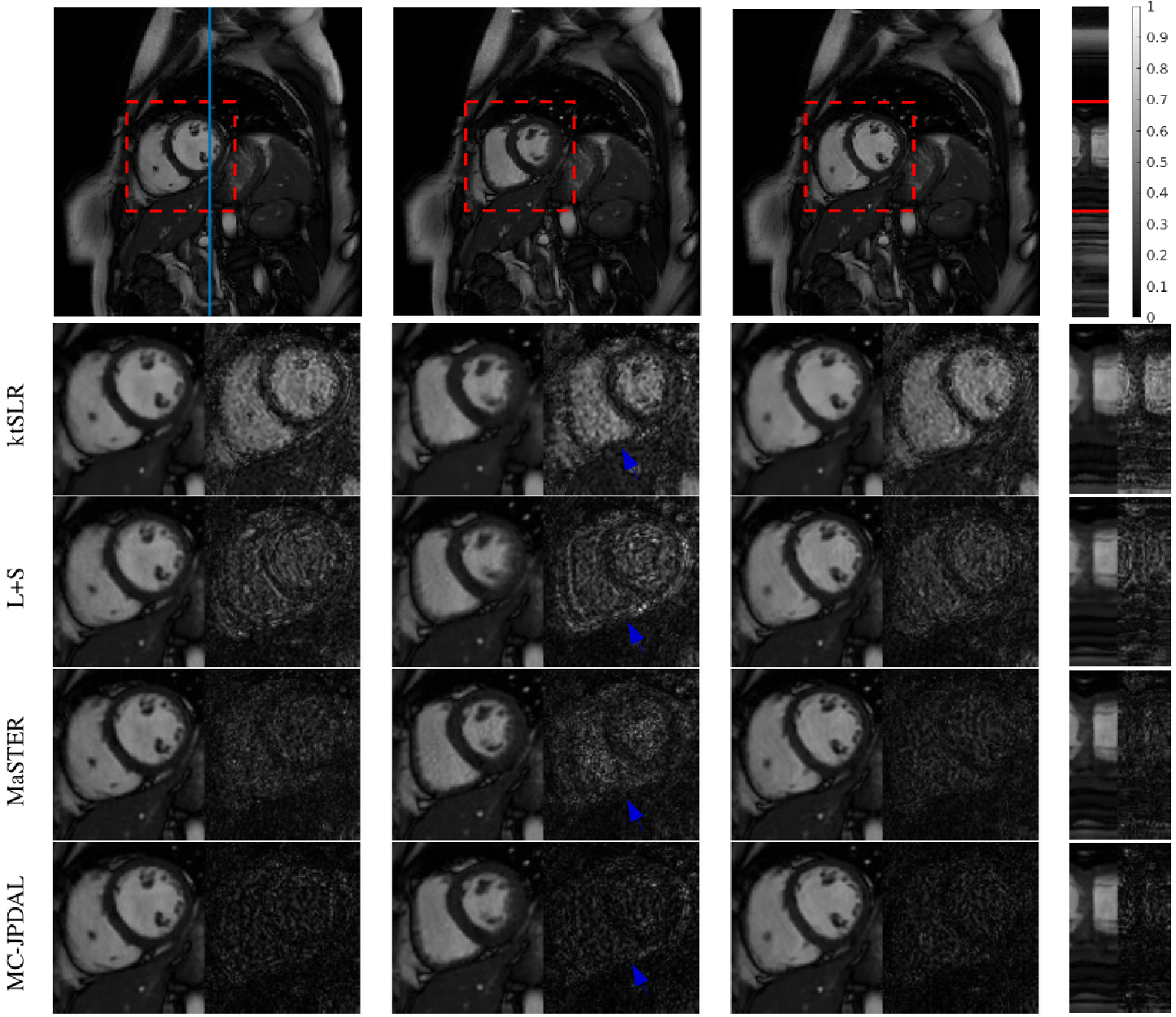}
\caption{Reconstruction of cardiac cine MRI scan using different algorithms: frame 3, 16 and 27 and the temporal profile (left to right). Top row: fully sampled MRI sequence with ROI contoured using red dashed rectangle and the location of the extracted temporal profile indicated using blue vertical line. Bottom rows: zoomed spatial ROI of the reconstructed MRI scans using ktSLR, L+S, MaSTER and the proposed MC-JPDAL.}
\label{fig_Car_MRI}
\end{center}
\end{figure*}

The reconstruction results are displayed in Fig. \ref{fig_Car_MRI}. The 1st row shows the fully sampled cardiac cine data at temporal frames 3, 16 and 27 and the temporal profile in $y$-$t$ space (from left to right). The ROIs are contoured by a red dashed rectangle. The location where the temporal profile extracted is indicated using a blue vertical line. From 2nd to 5th rows, the enlarged ROIs and their corresponding difference images ($\bff - \hat{\bff}$) of the reconstructed MRI frames using algorithms ktSLR, L+S, MaSTER and MC-PDAL are displayed. \tbme{Visually, the proposed MC-JPDAL outperforms the others since the reconstructed frames with the proposed algorithm are darker in terms of the magnitude of the difference images.} 

\begin{figure*}[!h]
\begin{center}
\includegraphics[width=0.95\linewidth]{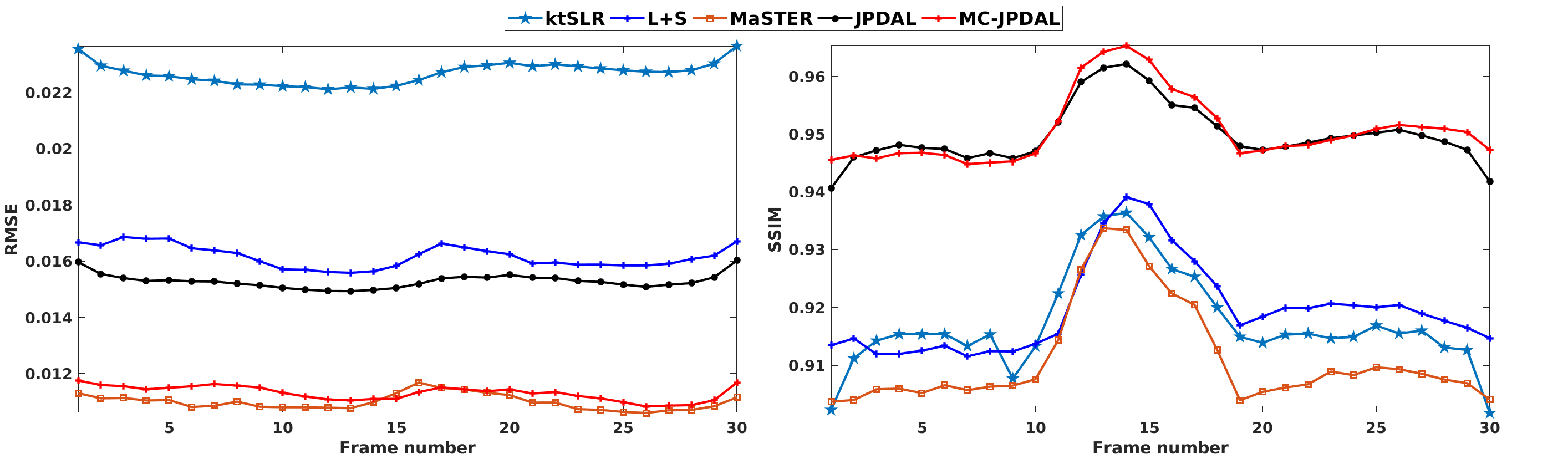}
\caption{Quantitative comparison of cardiac cine MRI sequences using the algorithms: ktSLR, L+S, MaSTER, \tbme{the proposed JPDAL and MC-JPDAL}. Left: RMSEs over the whole image; Right: SSIMs over the whole image.}
\label{fig_Car_curve}
\end{center}
\end{figure*}

Fig. \ref{fig_Car_curve} shows the quantitative measurements RMSE (left) and SSIM (right) calculated over the whole MRI frames. \tbme{The proposed algorithm MC-JPDAL outperforms the algorithms ktSLR, L+S and MaSTER in terms of the SSIM, which is consistent with the visual inspection.} The algorithms MC-JPDAL and MaSTER have comparable performance in terms of RMSE, which are superior the algorithms L+S and ktSLR. \tbme{The proposed MC-JPDAL also improves the image reconstruction quality compared with JPDAL in terms of RMSE and SSIM.}


\subsection{Two-chamber cardiac cine data}
The two-chamber cine MRI sequences were acquired using a Philips Intera 1.5T scanner with a 5-element cardiac synergy coil and a balanced fast field echo study-state free precession sequence. More details on the scan parameters can be found in \cite{Asif2013}. The sensitivity maps were estimated in advance. In this experiment, a 2D Cartesian down-sampling pattern with a fully sampled low-frequency region and a randomly sampled high-frequency region. The down-sampling/reduction factor was 10. \tbme{After comparing different priors for the reconstruction of the cardiac cine MRI sequences, the prior for this dataset is the combination of sparsity and TV prior (denoted as ``$\ell_1$+tv") in the proposed algorithms for further comparison.}

\begin{figure*}[!h]
\begin{center}
\includegraphics[width=0.95\linewidth]{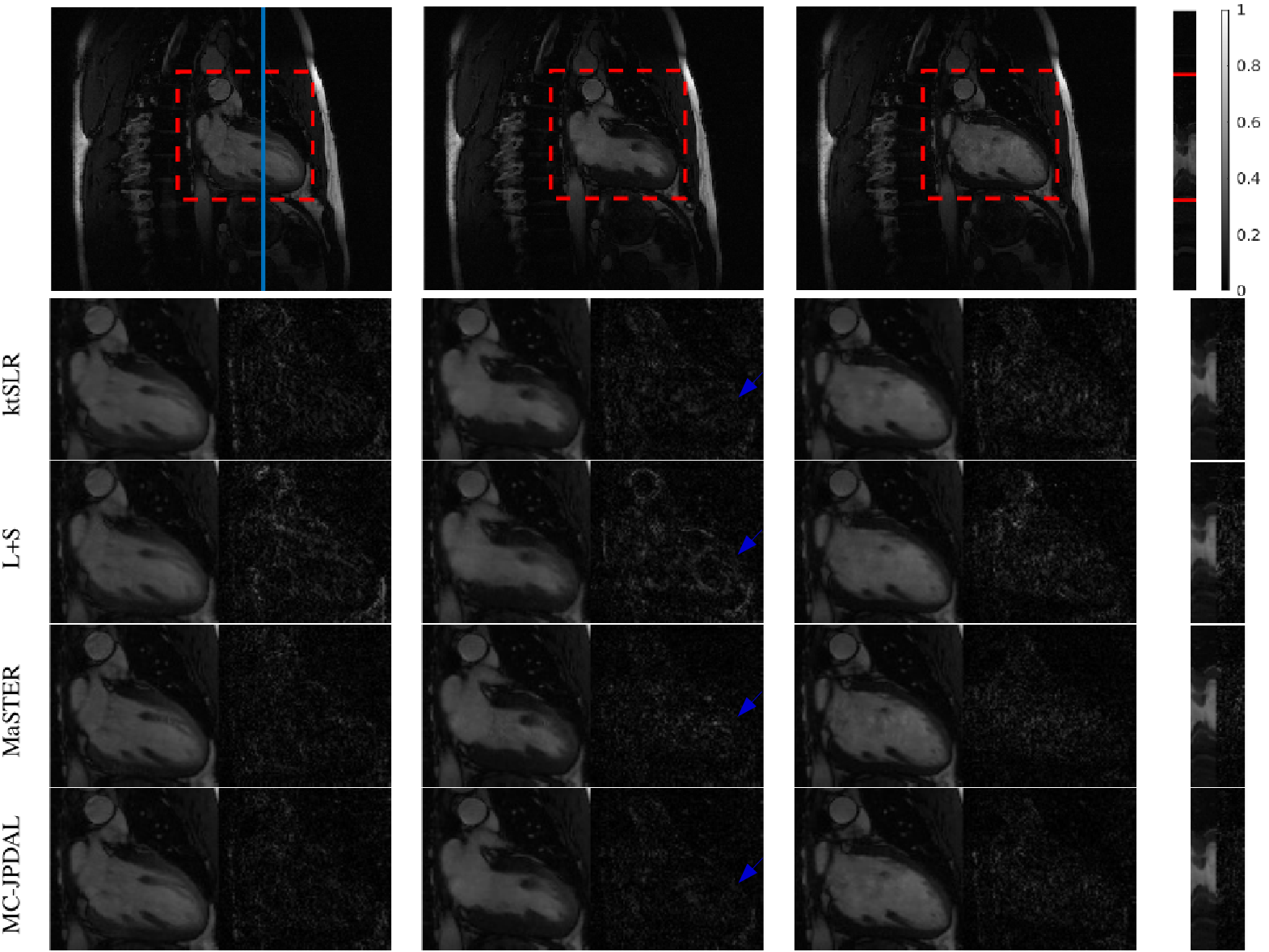}
\caption{Reconstruction of the two-chamber MRI scan using different algorithms: frames 3, 10, 14 and the temporal profile (left to right). Top row: fully sampled MRI sequence with ROI contoured using red dashed rectangle and the location of the extracted temporal profile indicated using blue vertical line. Bottom rows: zoomed spatial ROI of the reconstructed MRI scans using ktSLR, L+S, MaSTER and the proposed MC-JPDAL.}
\label{fig_TC_MRI}
\end{center}
\end{figure*}

Fig. \ref{fig_TC_MRI} illustrates the comparison of the reconstruction results using algorithms ktSLR, L+S, MaSTER and the proposed MC-JPDAL. The top row shows the frames 3, 10 and 14 out of 16 frames, constructed from fully sampled k-space data and the temporal profile in $y$-$t$ space (from left to right). The ROIs are contoured by a red dashed rectangle. The location where the temporal profile extracted is indicated using a blue vertical line. From 2nd to 5th rows, the enlarged ROIs and their corresponding difference images ($\bff - \hat{\bff}$) extracted from the reconstructed MRI sequences using ktSLR, L+S, MaSTER and the proposed MC-JPDAL are displayed. \tbme{In terms of the magnitude of the difference images, the proposed MC-JPDAL outperforms the others.} 

\begin{figure*}[!h]
\begin{center}
\includegraphics[width=0.95\linewidth]{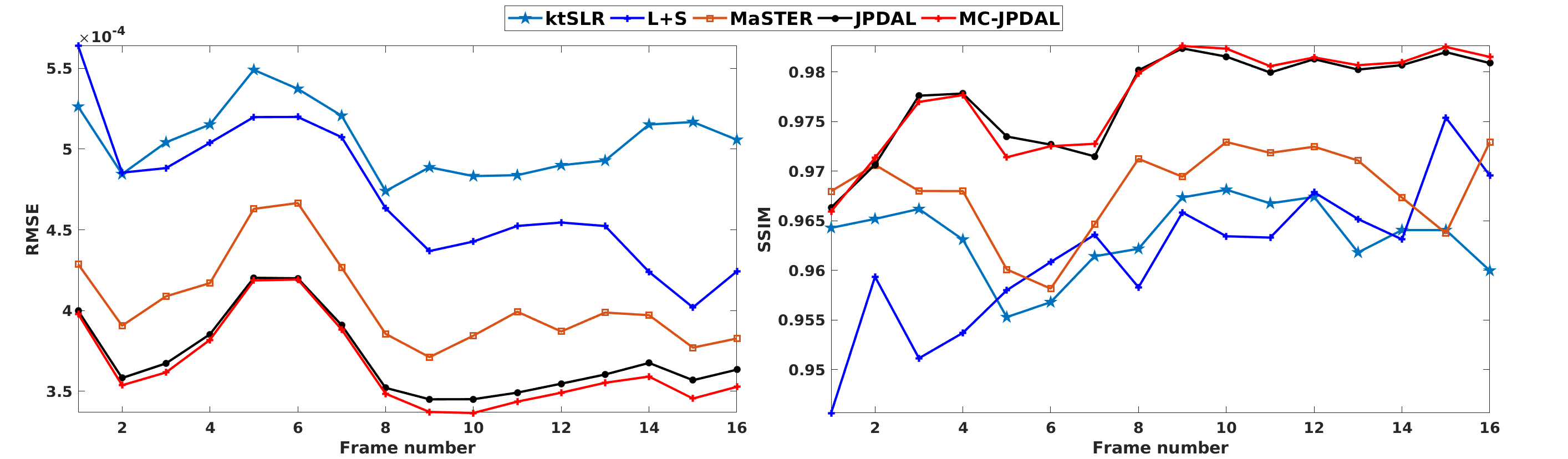}
\caption{Quantitative comparison of the two-chamber MRI sequences using the algorithms: ktSLR, L+S, MaSTER, \tbme{the proposed JPDAL and MC-JPDAL}. Left: RMSEs over the whole image; Right: SSIMs over the whole image.}
\label{fig_TC_SER}
\end{center}
\end{figure*}

Fig. \ref{fig_TC_SER} shows the quantitative comparison in terms of RMSE and SSIM calculated over the entire MRI sequences using the algorithms ktSLR, L+S, MaSTER, \tbme{JPAL and} MC-JPDAL. The proposed algorithms JPDAL and MC-JPDAL outperforms the others in terms of RMSE and SSIM. \tbme{We also observe that MC-JPDAL improves the reconstruction quality compared with JPDAL in terms of RMSE and SSIM.}

\begin{table}[!h]
\begin{center}
\caption{\tbme{Computational time (min) acquired with different methods for the three datasets}}
\begin{tabular}{|l||*{5}{c|}}\hline
\backslashbox[6mm]{Dataset}{Method}
&\makebox[2em]{ktSLR}&\makebox[2em]{L+S}&\makebox[3em]{MaSTER}&\makebox[2em]{JPDAL}&\makebox[3.5em]{MC-JDPAL}\\\hline\hline
Coronal Lung & 9.14 & 0.13 & 17.99 & 18.44 & 17.61\\\hline
Short-axis cardiac & 15.93& 0.35 & 58.85 & 57.94 & 39.05\\\hline
Two-chamber cardiac &  34.80 & 32.68 & 26.70 & 49.37 &36.40\\\hline
\end{tabular}
\label{tab_Time}
\end{center}
\end{table}
\tbme{Table \ref{tab_Time} summarizes the computational time for the three datasets in this section, where L+S outperforms the others in terms of computational time for the first and second datasets. We also note that the proposed algorithm MC-JPDAL is able to improve the image reconstruction quality of JPDAL without further computational burden.}

\zhao{Compared with other DMRI reconstruction algorithms, the proposed algorithm estimate the motion vectors and the image sequence jointly, which is one of the main contributions of this work. It is also interesting to note that both forward and backward motion patterns were considered for MC in MaSTER. The image reconstruction performance of the proposed method is comparable to MaSTER with only the forward motion. } 
%Although we only considered forward motion for the MC of DMRI reconstruction refinement, we have achieved state-of-the-art DMRI image reconstruction performance. Moreover, according to the comparison between JPDAL and MC-JPDAL, we notice that the joint variable estimation stage in the proposed algorithm plays a dominant role for the performance improvement of DMRI reconstruction.}
\section{Conclusions}
This paper proposed a novel framework to reconstruct DMRI using motion compensation, which \zhao{alternates between} two stages. One is to jointly estimate the DMRI frames and the motion vectors by combining the intensity based optical flow constraint with the compressed sensing framework, \zhao{which is one of the main contribution of the proposed MC-JPDAL}. \rev{Then, the estimated motion vectors are employed to refine the reconstructed DMRI sequence through motion compensation. By employing the coarse-to-fine multiscale strategy, the motion vectors can be estimated at different resolution scales.} The formulated problem is addressed using a primal dual algorithm with linesearch. In addition, the proposed scheme is able to deal with a wide class of image priors for DMRI reconstruction. We demonstrated that the proposed algorithm can obtain state-of-the-art DMRI reconstruction performance without necessarily to be the global minimum. 

%Future work includes \tbme{considering both forward and backward motion patterns for motion estimation and} the extension of the proposed algorithm to 3D+t DMRI data.

\section*{Acknowledgements}
This work has been supported by NIH grant R01CA188300. 

\bibliographystyle{IEEEtran}
\bibliography{strings_all_ref,bibli_Joint}

\begin{thebibliography}{10}
\providecommand{\url}[1]{#1}
\csname url@samestyle\endcsname
\providecommand{\newblock}{\relax}
\providecommand{\bibinfo}[2]{#2}
\providecommand{\BIBentrySTDinterwordspacing}{\spaceskip=0pt\relax}
\providecommand{\BIBentryALTinterwordstretchfactor}{4}
\providecommand{\BIBentryALTinterwordspacing}{\spaceskip=\fontdimen2\font plus
\BIBentryALTinterwordstretchfactor\fontdimen3\font minus
  \fontdimen4\font\relax}
\providecommand{\BIBforeignlanguage}[2]{{%
\expandafter\ifx\csname l@#1\endcsname\relax
\typeout{** WARNING: IEEEtran.bst: No hyphenation pattern has been}%
\typeout{** loaded for the language `#1'. Using the pattern for}%
\typeout{** the default language instead.}%
\else
\language=\csname l@#1\endcsname
\fi
#2}}
\providecommand{\BIBdecl}{\relax}
\BIBdecl

\bibitem{Mansfield1977}
\BIBentryALTinterwordspacing
P.~Mansfield, ``{Multi-planar image formation using NMR spin echoes},''
  \emph{Journal of Physics C: Solid State Physics}, vol.~10, no.~3, p. L55,
  1977. [Online]. Available: \url{http://stacks.iop.org/0022-3719/10/i=3/a=004}
\BIBentrySTDinterwordspacing

\bibitem{HAASE2011}
\BIBentryALTinterwordspacing
A.~Haase \emph{et~al.}, ``{FLASH imaging: Rapid NMR imaging using low
  flip-angle pulses},'' \emph{J. Magn. Reson.}, vol. 213, no.~2, pp. 533 --
  541, 2011. [Online]. Available:
  \url{http://www.sciencedirect.com/science/article/pii/S1090780711003338}
\BIBentrySTDinterwordspacing

\bibitem{Tsao2012}
J.~Tsao and S.~Kozerke, ``{MRI temporal acceleration techniques},'' \emph{J.
  Magn. Reson. Imaging}, vol.~36, no.~3, pp. 543--560, 2012.

\bibitem{Lustig2007}
M.~Lustig \emph{et~al.}, ``{Sparse MRI: The application of compressed sensing
  for rapid MR imaging},'' \emph{Magn. Reson. Med.}, vol.~58, no.~6, pp.
  1182--1195, 2007.

\bibitem{LustigMichaelandDonohoDavidLandSantosJuanMandPauly2008}
------, ``{Compressed sensing MRI},'' \emph{IEEE Sig. Process. Mag.}, vol.~25,
  no.~2, pp. 72--82, 2008.

\bibitem{Jung2009}
H.~Jung \emph{et~al.}, ``{K-t FOCUSS: A general compressed sensing framework
  for high resolution dynamic MRI},'' \emph{Magn. Reson. Med.}, vol.~61, no.~1,
  pp. 103--116, 2009.

\bibitem{Knoll2012}
F.~Knoll \emph{et~al.}, ``Parallel imaging with nonlinear reconstruction using
  variational penalties,'' \emph{Magn. Reson. Med.}, vol.~67, no.~1, pp.
  34--41, 2012.

\bibitem{Liang2007}
Z.~P. Liang, ``Spatiotemporal imaging with partially separable functions,'' in
  \emph{Joint Meeting of the 6th International Symposium on Noninvasive
  Functional Source Imaging of the Brain and Heart and the International
  Conference on Functional Biomedical Imaging (NFSI-ICFBI)}, Hangzhou, China,
  Oct 2007.

\bibitem{Trzasko2011ISMRM}
J.~Trzasko and A.~Manduca, ``Local versus global low-rank promotion in dynamic
  mri series reconstruction,'' in \emph{Proc. Annual Meeting of ISMRM},
  Qu\'ebec, Canada, 2011.

\bibitem{Miao2016}
\BIBentryALTinterwordspacing
X.~Miao \emph{et~al.}, ``{Accelerated cardiac cine MRI using locally low rank
  and finite difference constraints},'' \emph{Magn. Reson. Imaging}, vol.~34,
  no.~6, pp. 707--714, 2016. [Online]. Available:
  \url{http://dx.doi.org/10.1016/j.mri.2016.03.007}
\BIBentrySTDinterwordspacing

\bibitem{Lingala2011a}
S.~G. Lingala \emph{et~al.}, ``{Accelerated dynamic MRI exploiting sparsity and
  low-rank structure: k-t SLR},'' \emph{IEEE Trans. Med. Imag.}, vol.~30,
  no.~5, pp. 1042--1054, 2011.

\bibitem{Majumdar2015}
\BIBentryALTinterwordspacing
A.~Majumdar, ``{Real-time Dynamic MRI Reconstruction using Stacked Denoising
  Autoencoder},'' 2015. [Online]. Available:
  \url{http://arxiv.org/abs/1503.06383}
\BIBentrySTDinterwordspacing

\bibitem{Majumdar2012}
A.~Majumdar \emph{et~al.}, ``{Compressed sensing based real-time dynamic MRI
  reconstruction},'' \emph{IEEE Trans. Med. Imag.}, vol.~31, no.~12, pp.
  2253--2266, 2012.

\bibitem{Asif2013}
M.~S. Asif \emph{et~al.}, ``Motion-adaptive spatio-temporal regularization for
  accelerated dynamic {MRI},'' \emph{Magn. Reson. Med.}, vol.~70, pp. 800--812,
  2013.

\bibitem{Usman2013}
M.~Usman \emph{et~al.}, ``Motion corrected compressed sensing for
  free-breathing dynamic cardiac {MRI},'' \emph{Magn. Reson. Med.}, vol.~70,
  pp. 504--516, 2013.

\bibitem{Otazo2015}
R.~Otazo \emph{et~al.}, ``{Low-rank plus sparse matrix decomposition for
  accelerated dynamic MRI with separation of background and dynamic
  components},'' \emph{Magn. Reson. Med.}, vol.~73, no.~3, pp. 1125--1136,
  2015.

\bibitem{Tremoulheac2014}
B.~Tremoulheac \emph{et~al.}, ``{Dynamic MR image reconstruction-separation
  from undersampled (k,t)-Space via low-rank plus sparse prior},'' \emph{IEEE
  Trans. Med. Imag.}, vol.~33, no.~8, pp. 1689--1701, 2014.

\bibitem{Lingala2015}
S.~G. Lingala \emph{et~al.}, ``{( DC-CS ): A Novel Framework for Accelerated
  Dynamic MRI},'' \emph{IEEE Trans. Med. Imag.}, vol.~34, no.~1, pp. 72--85,
  2015.

\bibitem{Cordero-Grande2016}
J.~{Royuela-del-Val} \emph{et~al.}, ``Nonrigid groupwise registration for
  motion estimation and compensation in compressed sensing reconstruction of
  breath-hold cardiac cine {MRI},'' \emph{Magn. Reson. Med.}, vol.~75, pp.
  1525--1536, 2016.

\bibitem{Cordero-Grande2017}
L.~Cordero-Grande \emph{et~al.}, ``Three-dimensional motion corrected
  sensitivity encoding reconstruction for multi-shot multi-slice {MRI}:
  Application to neonatal brain imaging,'' \emph{Magn. Reson. Med.}, 2017.

\bibitem{Prieto2007}
C.~Prieto \emph{et~al.}, ``Reconstruction of undersampled dynamic images by
  modeling the motion of object elements,'' \emph{Magn. Reson. Med.}, vol.~57,
  pp. 939--949, 2007.

\bibitem{Jung2010}
H.~Jung and J.~C. Ye, ``{Motion estimated and compensated compressed sensing
  dynamic magnetic resonance imaging: What we can learn from video compression
  techniques},'' \emph{Int. J. Imaging Syst. Technol.}, vol.~20, no.~2, pp.
  81--98, 2010.

\bibitem{Feng2016}
L.~Feng \emph{et~al.}, ``{XD-GRASP: Golden-angle radial MRI with reconstruction
  of extra motion-state dimensions using compressed sensing},'' \emph{Magn.
  Reson. Med.}, vol.~75, no.~2, pp. 775--788, 2016.

\bibitem{MRM2017Rank}
C.~M. Rank \emph{et~al.}, ``4d respiratory motion-compensated image
  reconstruction of free-breathing radial mr data with very high
  undersampling,'' \emph{Magn. Reson. Med.}, vol.~77, no.~3, pp. 1170--1183,
  2017.

\bibitem{Malitsky2016}
\BIBentryALTinterwordspacing
Y.~Malitsky and T.~Pock, ``{A first-order primal-dual algorithm with
  linesearch},'' pp. 1--24, 2016. [Online]. Available:
  \url{http://arxiv.org/abs/1608.08883}
\BIBentrySTDinterwordspacing

\bibitem{Suhling2005}
M.~S{\"{u}}hling \emph{et~al.}, ``{Myocardial motion analysis from B-mode
  echocardiograms},'' \emph{IEEE Trans. Image Process.}, vol.~14, no.~4, pp.
  525--536, 2005.

\bibitem{nzhao20158SPIE}
N.~Zhao \emph{et~al.}, ``{Coupling reconstruction and motion estimation for
  dynamic MRI through optical flow constraint},'' \emph{Proceedings SPIE
  Medical Imaging}, vol. 10574, 2018.

\bibitem{MRI_handbook2004}
M.~A. Bernstein \emph{et~al.}, \emph{Handbook of {MRI} Pulse Sequences}.\hskip
  1em plus 0.5em minus 0.4em\relax Elsevier Academic Press, 2004.

\bibitem{SunCVPR2010}
D.~Sun \emph{et~al.}, ``Secrets of optical flow estimation and their
  principles,'' in \emph{Proc. IEEE Conference on Computer Vision and Pattern
  Recognition (CVPR)}, San Francisco, CA, USA, 2010.

\bibitem{Altunbasak2003}
Y.~Altunbasak \emph{et~al.}, ``{A fast parametric motion estimation algorithm
  with illumination and lens distortion correction},'' \emph{IEEE Trans. Image
  Process.}, vol.~12, no.~4, pp. 395--408, 2003.

\bibitem{Alessandrini2013}
M.~Alessandrini \emph{et~al.}, ``{Myocardial Motion Estimation from Medical
  Images Using the Monogenic Signal},'' \emph{IEEE Trans. Image Process.},
  vol.~22, no.~3, pp. 1084--1095, 2013.

\bibitem{OPT-003}
\BIBentryALTinterwordspacing
N.~Parikh and S.~Boyd, ``Proximal algorithms,'' \emph{Foundations and Trends in
  Optimization}, vol.~1, no.~3, pp. 127--239, 2014. [Online]. Available:
  \url{http://dx.doi.org/10.1561/2400000003}
\BIBentrySTDinterwordspacing

\bibitem{pock2009algorithm}
T.~Pock \emph{et~al.}, ``An algorithm for minimizing the mumford-shah
  functional,'' in \emph{Computer Vision, 2009 IEEE 12th International
  Conference on}.\hskip 1em plus 0.5em minus 0.4em\relax IEEE, 2009, pp.
  1133--1140.

\bibitem{Chambolle2011}
A.~Chambolle and T.~Pock, ``{A first-order primal-dual algorithm for convex
  problems with applications to imaging},'' \emph{J. Math. Imag. Vision},
  vol.~40, no.~1, pp. 120--145, 2011.

\bibitem{Komodakis2015}
N.~Komodakis and J.~C. Pesquet, ``{Playing with duality: An overview of recent
  primal-dual approaches for solving large-scale optimization problems},''
  \emph{IEEE Sig. Process. Mag.}, vol.~32, no.~6, pp. 31--54, 2015.

\bibitem{esser2009general}
E.~Esser \emph{et~al.}, ``A general framework for a class of first order
  primal-dual algorithms for tv minimization,'' \emph{Ucla Cam Report}, pp.
  09--67, 2009.

\bibitem{Wang2004}
Z.~Wang \emph{et~al.}, ``Image quality assessment: From error visibility to
  structural similarity,'' \emph{IEEE Trans. Image Process.}, vol.~13, no.~4,
  pp. 600--612, 2004.

\bibitem{Feng2014}
L.~Feng \emph{et~al.}, ``Golden-angle radial sparse parallel {MRI} :
  Combination of compressed sensing , parallel imaging , and golden-angle
  radial sampling for fast and flexible dynamic volumetric {MRI},'' \emph{Magn.
  Reson. Med.}, vol.~72, pp. 707--717, 2014.

\bibitem{Caballero2014}
J.~Caballero \emph{et~al.}, ``Dictionary learning and time sparsity for dynamic
  mr data reconstruction,'' \emph{IEEE Trans. Med. Imag.}, vol.~33, no.~4, pp.
  979--994, 2014.

\end{thebibliography}
\end{document}